\documentclass{article}
\usepackage{ijcai22}

\usepackage{times}
\usepackage{soul}
\usepackage{url}
\usepackage[hidelinks]{hyperref}
\usepackage[utf8]{inputenc}
\usepackage[small]{caption}
\usepackage{graphicx}
\usepackage{amsmath}
\usepackage{amsthm}
\usepackage{booktabs}
\usepackage{algorithm}
\usepackage{algorithmic}

\usepackage{amsfonts}       
\usepackage{nicefrac}       
\usepackage{microtype}      
\usepackage{xcolor}         
\usepackage{verbatim}

\usepackage{amssymb}

\usepackage{wrapfig}

\usepackage{subfigure}
\usepackage{array}
\usepackage{tabu}

\newtheorem{thm}{Theorem}

\def\ie{\emph{i.e.}}

\def\etal{\emph{et al.}}

\title{Confidence Dimension for Deep Learning based on Hoeffding Inequality and Relative Evaluation}

\author{
Runqi Wang$^1$
\and
Linlin Yang$^2$\and
Baochang Zhang*$^{1}$\and
Wentao Zhu$^3$\and
David Doermann$^4$\and
Guodong Guo$^5$
\affiliations
$^1$Beihang University
$^2$University of Bonn
$^3$Amazon
$^4$University at Buffalo
$^4$Baidu
}

\begin{document}

\maketitle

\begin{abstract}
Research on the generalization ability of deep neural networks (DNNs) has recently attracted a great deal of attention. However, due to their complex architectures and large numbers of parameters, measuring the generalization ability of specific DNN models remains an open challenge. In this paper, we propose to use multiple factors to measure and rank the relative generalization of DNNs based on a new concept of \emph{confidence dimension} ($CD$). Furthermore, we provide a feasible framework in our $CD$ to theoretically calculate the upper bound of generalization based on the conventional Vapnik-Chervonenk dimension (VC-dimension) and Hoeffding's inequality.
Experimental results on image classification and object detection demonstrate that our $CD$ can reflect the relative generalization ability for different DNNs.
In addition to full-precision DNNs, we also analyze the generalization ability of binary neural networks (BNNs), whose generalization ability remains an unsolved problem. Our $CD$ yields a consistent and reliable measure and ranking for both full-precision DNNs and BNNs on all the tasks.
\end{abstract}

\section{Introduction}
\label{introduction}
Deep neural networks (DNNs) contribute significantly to many computer vision tasks over the past decade. However, most DNNs are heavily over-parameterized, which means that the models can achieve near-perfect performance on the training data but have a performance gap on the test data.
Measuring the generalization ability is complex. This raises a fundamental question: can we fairly and objectively evaluate the generalization ability of a deep learning model?

Theoretical research on the image classification capacity of shallow models starts with Vapnik-Chervonenkis Dimension (VC-dimension)~\cite{vapnik1971uniform}. 
Taking the data distribution into account, Bartlett~\etal~\cite{bartlett2002rademacher} further proposes the Rademacher complexity, a stricter generalization error bound. 

The training of DNNs is more complicated and involves solving nonconvex optimization problems in a very high-dimensional space. Moreover, it is difficult to find the global minimum of DNNs analytically. Previous works~\cite{dziugaite2017computing} have shown that there is a degree of correlation between the optimizer and the generalization ability of DNNs. 
Inspired by this, Fort~\etal~\cite{fort2019stiffness} proposes a gradient alignment measure to understand the generalization. 
However, the optimizer is not the only factor that affects the generalization ability. For example, training different network architectures with the same optimizer may lead to other generalization errors. 
Furthermore, batch size and regularization techniques such as dropout~\cite{srivastava2014dropout}, weight decay, and early stopping also influence the generalization ability. Those regularization techniques lower the model's complexity and therefore affect the generalization.

Based on these observations, the generalization of DNNs should consider both the network architecture and the optimization. Due to finite sampling set, traditional VC-dimension can only produce quite loose generalization bounds, which are unsuitable to describe DNNs with more parameters than training samples~\cite{valle2020generalization,zhang2016understanding}. Tighter generalization bounds~\cite{neyshabur2018towards,valle2020generalization} have been proposed to understand the generalization ability of deep models better. 
However, these generalization bounds are still impractical for real applications because of their restrictive assumptions. For example, Valle~\etal~\cite{valle2018deep} ignores the effects of different training ``tricks'' on the generalization. Furthermore, Neyshabur~\etal~\cite{neyshabur2018towards} limits the architecture to two-layer networks, and Valle~\etal~\cite{valle2020generalization} restricts the studied image classification tasks to binary problems.
Moreover, previous methods based on VC-dimension often offer an unstable explanation and occasionally show trends opposite the actual error. Previous bound predictions increase with the increasing training set size, while the measured error decreases~\cite{nagarajan2019uniform}.  As a result, for real applications only when a finite training sample set is available for the VC-dimension calculation, a theoretical upper bound for generalization will provide a guidance to analyze DNNs from a pragmatic perspective.

\begin{figure*}[t]
\begin{center}
\includegraphics[width=0.7\linewidth]{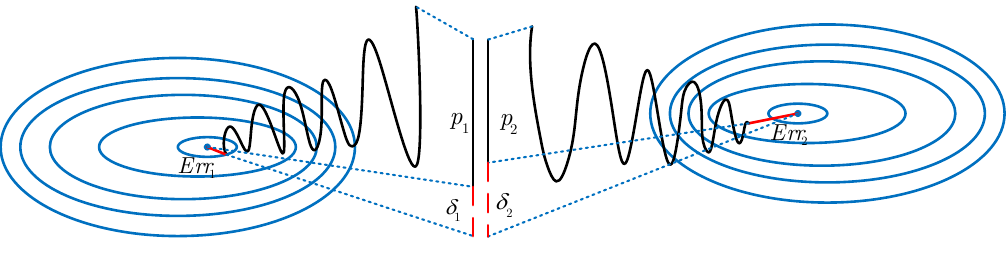}
\caption{The illustration of Confidence Dimension. The traditional VC-dimension estimation ignores the process of optimization, which lead to different measure results ($p_1$ and $p_2$) for the  VC-dimension of the same model.  To address the issue, we theoretically calculate the upper bound of the VC-dimension based on a correction term ($\delta_1$ and $\delta_2$) which is positively related to the training error for a stable generalization estimation. }
\label{moti}
\end{center}
\vspace{-0.7cm}
\end{figure*}

So far, there has not been any effective method to measure generalization. In this paper, we propose a new concept of \emph{confidence dimension} ($CD$) based on VC-dimension and Hoeffding's inequality to fairly evaluate the generalization ability of DNNs, by considering both the network architecture and the optimization process in a unified framework, as shown in Fig.~\ref{moti}. Many factors are not taken into account in the VC-dimension~\cite{zhang2016understanding}. Now we consider inputs and process of optimization as a correction term to obtain the upper bound of generalization, so as to ensure that the generalization of models are compared within a unified and relatively completed framework. It also can be used to measure the generalization of models on a unified scale under different experimental conditions. We investigate the generalization evaluation based on a relative assessment of different DNN models, i.e., focus on the $CD$ ranking consistency of DNN models across different situations, which is more practical than existing methods~\cite{bartlett2002rademacher,vapnik1971uniform}. We take two steps to establish an upper bound for generalization, leading to our \emph{confidence dimension} ($CD$), and provide explanatory power for DNNs. 
First,  DNN models are trained on randomly and accurately labeled data to calculate the VC-dimension.  Then, based on Hoeffding's inequality, we calculate the corrected term of generalization by considering the optimization process~\cite{dziugaite2020search} into calculating the $CD$. According to the law of large numbers, the $CD$ will become stable by performing large numbers of experiments. 

We demonstrate the rationale of our method theoretically and experimentally. We first provide a theoretical derivation of the upper bound of generalization based on the VC-dimension and Hoeffding's inequality. Then, we further extensively validate the generalization bound $CD$ by cross-tasks, including image classification and object detection.  Moreover, different from existing works analyzing the generalization ability only for full-precision models, we also evaluate the generalization ability for binary neural networks (BNNs). BNNs~\cite{gu2019projection,hubara2016binarized,liu2020reactnet} are among the most compressed deep models and are worth exploring their generalization performance, which quantize weights and features to single bits and have
attracted intense interest for their excellent computation acceleration and model compression. 
The main contributions of this paper are as follows:
\begin{itemize}
    \item We propose a \emph{confidence dimension} ($CD$) to measure the relative generalization ability of deep models and provide a new approach for generalization prediction.
    \item We show that theoretically, our $CD$ is the upper bound of generalization based Hoeffding's inequality, which is more feasible than conventional measures when only a data-driven method is available to measure the generalization ability of DNNs.
    \item  We extensively validate the performance of our $CD$ by cross-tasks, including image classification and object detection, BNNs  and full-precision DNNs, which are consistent with the theoretical results. The $CD$ measure of a DNN is stable for various tasks and is potentially a useful tool to evaluate the generation for DNNs. 
\end{itemize}

\section{Related Work}
\label{sec:related}
The VC-dimension~\cite{vapnik1971uniform} provides a general measure of the complexity of traditional models and reflects their learning ability. It is closely related to the generalization performance of the underlying models. To guarantee good generalization performance, we expect a novelty dimension to give a tight bound on the expected error. Unfortunately, the VC-dimension has practical limitations when applied to the generalization of DNNs. Contrary to VC-dimension theory, decreasing the number of parameters in DNNs may achieve a better generalization performance. This is primarily because having more parameters than the number of training examples usually results in a loose generalization bound based on the VC-dimension~\cite{zhang2016understanding}.  Furthermore, different optimizers or regularization techniques have an influence on the optimization process and lower the model complexity, affecting the generalization~\cite{srivastava2014dropout}. Gintare~\etal~\cite{dziugaite2020search} argues that generalization measures should instead be evaluated within the framework of distributional robustness. Bengio~\etal~\cite{jiang2019fantastic} confirms that implicit regularization and optimizer influence the model in the same trend, i.e., the process of optimization is a comprehensive measurement of optimizer and regularization. 

Existing works mainly focus on estimating a tighter generalization bound to understand the generalization ability of DNN models better.
Sun~\etal~\cite{sun2016depth} expects the generalization error to be bounded by an empirical margin error plus the Rademacher Average term. Instead, works~\cite{valle2020generalization} prefer the PAC-Bayes bounds. Valle~\etal~\cite{valle2018deep} computes the PAC-Bayes generalization error bounds using the Gaussian process approximation of the prior over functions. Furthermore, Valle~\etal~\cite{valle2020generalization} proposes a marginal-likelihood PAC-Bayes bound under the assumption of a power-law asymptotic behavior with training set size.

Yet strong assumptions are still key bottlenecks preventing those generalization bounds from practical use. To evaluate the generalization of DNNs more flexibly, we propose a new measurement method that comprehensively considers the model's structure, optimization, and other training details like the training set size. In this case, we can compare the upper bound of the generalization ability between models in practical applications.

\section{Measuring Relative Generalization Ability}
\label{sec:method}
Our goal is to present a flexible framework to compare the generalization ability of different networks. Based on randomization tests also used in VC-dimension, we calculate the \emph{confidence dimension} ($CD$) to measure the models' generalization performance. In the following, we will introduce VC-dimension, $CD$, and then analyze $CD$ theoretically.

\subsection{VC-dimension Estimation}
\label{sec:vc}
Let $\mathcal{H}$ be a hypothetical space, and $\mathcal{X} = \left\{x_1,x_2,...,x_m\right\}$ be a sampling set with size $m$. Each hypothesis $h$ in $\mathcal{H}$ marks a sample in $\mathcal{X}$, and the result is expressed as:
\begin{equation}
    h|\mathcal{X} = \left\{h(x_1),h(x_2),...,h(x_m)\right\}.
\end{equation}
As the size of sample $m$ increases, the number of corresponding examples in $\mathcal{X}$ may also increase. For $m \in N$, the growth function is defined as:
\begin{equation}
    \Pi_\mathcal{H}(m) = \max_{{x_1,...,x_m}\subseteq \mathcal{X}} | \left\{h(x_1),h(x_2),...,h(x_m) | h\in \mathcal{H} \right\} |.
\end{equation}
The growth function $\Pi_\mathcal{H}(m)$ represents the maximum number of possible results that can be labeled with the hypothesis space $\mathcal{H}$ for $m$ examples. The greater the number of possible results that $\mathcal{H}$ can label for these examples, the stronger the expression ability of $\mathcal{H}$. In deep learning, the DNN is the hypothesis space $\mathcal{H}$.



Now let us double the number of sample sets $\mathcal{X} = \{x_1,...,x_m,x_{m+1},...,x_{2m}  \}$ and generate subsets $\mathcal{X}_1 = \{x_1,...,x_m\}$ and $\mathcal{X}_2 = \{x_{m+1},...,x_{2m} \} $. The risk of $\mathcal{X}$ in the deep learning model is defined as~\cite{vapnik1971uniform}:
\begin{equation}
    v(\mathcal{X}) =\frac{1}{2m} \sum_{i=0}^{2m} |y_i-f(x_i)|,
\end{equation}
where $y_i$ is the label of the dataset and $f(x_i)$ is the classification results of the model. Therefore, the risks of the two subsets are:
\begin{equation}
\begin{aligned}
    &v(\mathcal{X}_1) =\frac{1}{m} \sum_{i=0}^{m} |y_i-f(x_i)|, \\
    &v(\mathcal{X}_2) =\frac{1}{m} \sum_{i=m+1}^{2m} |y_i-f(x_i)|.\\
\end{aligned}
\end{equation}
According to previous work~\cite{vapnik1971uniform}, the difference between the two risks $v(\mathcal{X}_1)$ and $v(\mathcal{X}_2)$ is positively correlated with the sample set size $m$. 
\begin{equation}
     {p} = \sup\{v(\mathcal{X}_1)-v(\mathcal{X}_2)\} \propto m,
\end{equation}
$\sup$ denotes the upper limit, which  can be obtained by maximizing:
\begin{equation}
\begin{aligned}
&\frac{1}{m} \sum_{i=1}^{m}|y_i-f(x_i)| - \frac{1}{m}\sum_{i={m}+1}^{2m}|y_i-f(x_i)| \\
&=(1- \frac{1}{m}\sum_{i=1}^{m}|\tilde{y}_i-f(x_i)|) -  \frac{1}{m} \sum_{i={m}+1}^{2m}|{y_i}-f(x_i)|, \\
\end{aligned} \label{eq6a}
\end{equation}
\noindent which is equivalent to :
\begin{equation}
\begin{aligned}
 {p} = \inf \{( \frac{1}{m} \sum_{i=1}^{m}|\tilde{y}_i-f(x_i)| +  \frac{1}{m}\sum_{i={m}+1}^{2m}|y_i-f(x_i)|) \},
\end{aligned} \label{eq6b}
\end{equation}
where $\tilde{y}_i$ is the incorrect label of the dataset, $\inf$ denotes lower limit. When the sample set size $m$ is the same, VC-demision is proportional to $ {p}$. The result of $p$ will be normalized. There  exist $ {\zeta,\varepsilon}\in(0,\infty)$, such that the form of VC-dimension is shown below~\cite{vapnik1971uniform}:
\begin{equation}
\begin{aligned}
VC \propto \frac{\zeta}{e^{-\frac{m\varepsilon^2}{8}}} {p} .
\end{aligned} \label{eq8}
\end{equation} 
We can know that $ {p}$ is a measure of VC-dimension, which can be used to rank  VC-dimension of  different models. The minimum value in Eq.~\ref{eq6b} cannot be easily estimated because of  the huge complexity of deep learning models. We can only approximate the global minimum with a local minimum, which is  actually related to the used  optimizer. Furthermore, finding the maximum number of shattered samples still remains an open problem. As a result, we can only obtain an inaccurate generalization measure for DNNs by the method of VC-dimension.

\subsection{Confidence Dimension}
\label{sec:cd}
We consider multiple elements to define a new measure $CD$, which can provide a much more stable measure than VC-dimension. To this end, we calculate the correction term $\delta$ of generalization, which can be used to bound the generalization ability.  $\delta$   is estimated as:
\begin{equation}
    \delta=\alpha\sqrt{\frac{\ln (2+Err)}{m}},
\label{eq:u}
\end{equation}
where $Err$ denotes the training error for the sample set $\mathcal{X}$, $\alpha$ denotes the coefficient of correction terms, and $m$ is the size of $\mathcal{X}$.  We then define the  \emph{confidence dimension} ($CD$)  as:
\begin{equation}
    CD = \sup\{v(\mathcal{X}_1)-v(\mathcal{X}_2)\}+ \delta,
\label{eq:cd}
\end{equation}
in order to ensure $CD \in [0,1]$, $CD$ will be normalized. The advantages of our measure lie in that: 1) Our $CD$ can be more consistent than VC-dimension. Based on our theoretical investigation, we show that our measure is the upper bound of generalization based on the VC-dimension, which might have a tighter constraint; 2) Our $CD$ can effectively measure the relative generalization ability of different DNNs based on a finite sample size.

\begin{thm}\label{anti-probability}
The Confidence Dimension is an upper bound of the generalization ability of any model with a probability of $$1-\frac{1}{(2+Err)^4}$$
where $Err$ is training error.
\end{thm}

\noindent \textbf{Proof.}   We denote $\hat{p}_i$ as the normalized optimal value  for the  $i^{th}$ batch of independent and identically distributed samples, which is the infimum calculated by Eq.~\ref{eq6b}. We also define: 

$$\hat{p} = \frac{\sum \hat{p}_i}{m},$$
where $m$ is the number of input samples.  For easy proof our result,  we reasonably assume  $\hat{p}$  as  the probability measure of obtaining the upper bound of generalization.  
Due to the independence of $CD_i$ (in the $i^{th}$ batch), we can get Eq.~\ref{eq:theorem1} based on the  Hoeffding's  inequality~\cite{dubhashi2009concentration}:
\begin{equation}
    \begin{aligned}
    &\mathbb{E}[e^{\lambda CD}]= \mathbb{E}[e^{\frac{\lambda}{m} \sum_i[CD_i]}]\\
    &=\prod_i\mathbb{E}({e^{\frac{\lambda}{m} CD_i}})
    = \prod_i(\hat{p}_i e^\frac{\lambda}{m} + \hat{q}_i) \\
    &\leq  (\frac{\sum_i(\hat{p}_i e^\frac{\lambda}{m} + \hat{q}_i)}{m})^m
    =(\hat{p} e^\frac{\lambda}{m} +\hat{q})^m,\\
    \end{aligned}
    \label{eq:theorem1}
\end{equation}
where $\hat{q}_i=1-\hat{p}_i$ and $ \hat{q} = 1-\hat{p}$. Given Markov's inequality, we have:
\begin{equation}
    \begin{aligned}
    P[CD \geq \hat{p}+\delta]&=P[\frac{1}{m}\sum_i(CD_i - \hat{p}_i) \geq \delta] \\
    &=P[e^{ \frac{\lambda}{m}\sum_i(CD_i -\hat{p}_i)} \geq e^{\lambda \delta}]\\
    &\leq \frac{\mathbb{E}[e^{\frac{\lambda}{m} \sum_i(CD_i - \hat{p}_i)}]}{e^{\lambda \delta}}.
    \end{aligned}
    \label{eq:independent4global}
\end{equation}
We also have $1+x\leq e^x\leq 1+x+x^2,$ when $0 \leq |x| \leq 1$. We thus have $\mathbb{E}[e^{\frac{\lambda}{m}\sum_i(CD_i-\hat{p}_i)}]$ in Eq.~\ref{eq:independent4global}, which can be further approximated as:
 \begin{equation}
   \begin{aligned}
   &\mathbb{E}[e^{\frac{\lambda}{m}\sum_i(CD_i-\hat{p}_i)}]\\
   &=\prod_i \mathbb{E}[e^{\frac{\lambda}{m}(CD_i - \hat{p}_i)}] \\
   &\leq \prod_i \mathbb{E}[1+\frac{\lambda}{m}(CD_i - \hat{p}_i)
   +(\frac{\lambda}{m})^2(CD_i - \hat{p}_i)^2]\\
   &=\prod_i(1+(\frac{\lambda}{m})^2v_i^2)
   \leq e^{(\frac{\lambda}{m})^2 v^2},
   \end{aligned}
   \label{eq:denominator}
\end{equation}
\noindent where $v_i$ denotes the variance of $CD_i$, $v$ denotes the variance of $CD$. Combining Eq.~\ref{eq:independent4global} and Eq.~\ref{eq:denominator} gives $P[CD\geq \hat{p}+\delta] \leq \frac{e^{(\frac{\lambda}{m})^2 v^2}}{e^{\lambda \delta}}$. 
Since $\lambda$ is a non-negative constant and $CD\in [0,1]$,according to the root formula, it can be obtained by transforming values such that $P[CD \geq \hat{p}+\delta] \leq e^{-2m\delta^2}$ when $m\geq 8$. From the symmetry of the distribution, we have $P[CD < \hat{p}-\delta] \leq e^{-2m\delta^2}$. Finally, we get the inequality:
   \begin{equation}
       P[|CD - \hat{p}|\leq \delta] \geq 1- 2e^{-2m\delta^2},
       \label{eq:bound}
   \end{equation}

where $\hat{p}$ is is the normalized theoretical VC-dimension value. We guarantee that  $\delta$ decrease as the size of the input samples  increases. Therefore, we choose  $\alpha\sqrt{\frac{\ln{(2+Err)}}{m}}$ to be $\delta$ and have:

\begin{equation}
    \hat{p}-\alpha\sqrt{\frac{\ln{(2+Err)}}{m}}\leq CD \leq \hat{p}+\alpha\sqrt{\frac{\ln{(2+Err)}}{m}},
\end{equation}

which  shows that  the value of $CD$ as defined in Eq.~\ref{eq:cd} is the upper bound that the model can reach  with at least a probability of $1-\frac{1}{(2+Err)^4}~\cite{dubhashi2009concentration}$.

When $m$ is infinite, the value of the correction term will gradually decrease, and $\hat{p}$ will finally approach the $CD$. Eq.~\ref{eq:probability} shows that we can achieve a 0.988 probability when the training error gets to 1.
\begin{equation}
    1-\frac{1}{(2+Err)^4}=
    \begin{cases}
    0.938& \quad Err = 0\\
    0.988& \quad Err = 1.
\end{cases}
\label{eq:probability}
\end{equation}

By introducing a correction term into VC-dimension, we obtain a better estimation of the generalization ability. Experimentally, we also show that it can provide a relatively stable  measure of generalization ability for different DNNs.


\section{Experiments}
\label{sec:experiments}
We analyze the generalization ability for a variety of DNNs including BNNs based on $CD$. We also validate the performance of different object detectors.

\subsection{Implementation Details}
\label{sec:dataset}
\begin{figure*}[htp]
\begin{center}
\includegraphics[width=0.85\linewidth]{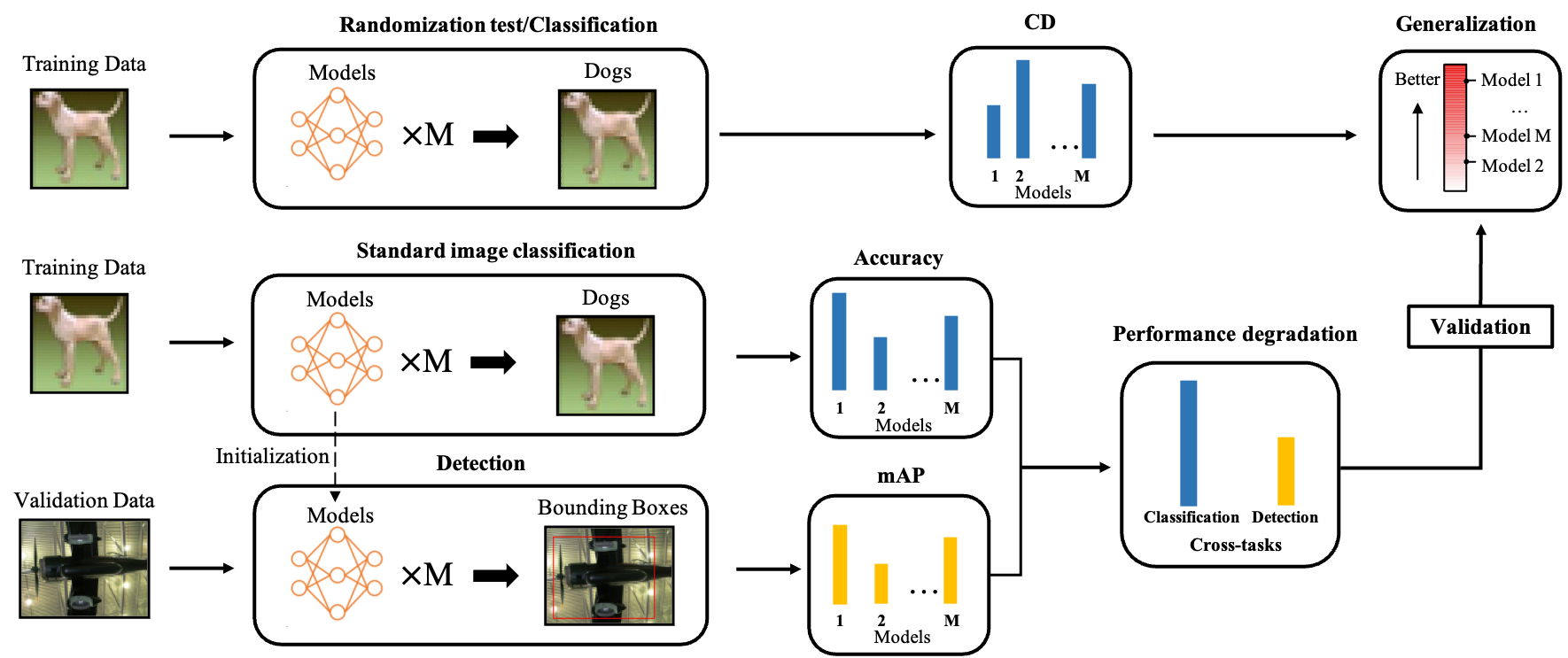}
\caption{The obtainment and validation of $CD$.  Based on the measure of VC-dimension ($p$), training set size, and training error, we calculate the $CD$ of DNNs via a image classification task (Randomization test). Since our proposed $CD$ is a general performance metric regardless of the task, we further validate $CD$ via the cross-tasks (standard image classification and object detection).} 
\label{fig:process}
\end{center}
\end{figure*}

\textbf{Cross-tasks:} Image classification and object detection are two fundamental computer vision tasks. We term traditional image classification as standard image classification, which involves predicting the category of one object in an image. As two different tasks (cross-tasks) are based on the same architecture,  we use the performance degradation  to validate the generalization ability of DNNs. Specifically, we calculate  $CD$ of DNNs via image classification and validate it via object detection as shown in Fig.~\ref{fig:process}. We demonstrate the consistency of $CD$ across two different tasks which better than VC-dimension.

\textbf{Randomization tests:}  Our methodology is based on a randomization test~\cite{zhang2016understanding}. A randomization test is a permutation test based on randomization from non-parametric statistics. We train several candidate architectures on a copy of the data where half of the correct labels are replaced with incorrect labels as described in Eq.~\ref{eq6b}. 
For brevity, we define two kinds of labels in our experiments: 1) \textbf{correct labels}: the original labels without modification, 2) \textbf{incorrect labels}: correct labels replaced with incorrect ones.

\textbf{Dataset:}
We modify CIFAR-10~\cite{krizhevsky2009learning} and ImageNet ILSVRC 2012~\cite{deng2009imagenet} for a more straightforward calculation of VC-dimension based on \cite{vapnik1971uniform}, which is a fundamental step to calculate the $CD$ of DNNs, then we validate the $CD$  on the PASCAL VOC dataset~\cite{everingham2015pascal} and MS COCO dataset~\cite{lin2014microsoft}. 
In order to test the effect of training set size on $CD$, we extract 2,5,7,10 classes from the CIFAR-10 dataset and 2,10,50 classes from the ImageNet dataset. We re-label these datasets with random half incorrect and correct labels as required by the randomization test method.

\textbf{BNNs:}
BNNs aim to compress the CNN models by quantizing the weights in the trained CNNs. We are the first to study the generalization ability of BNNs and estimate the performance bound of BNNs. We have selected two commonly used BNNs, PCNN~\cite{gu2019projection} and ReActNet~\cite{liu2020reactnet},  in our experiments. We choose PCNN in our experiments because it employs a new BP algorithm in the training process, which can be used to evaluate the generalization ability of different BNNs considering the optimization. ReActNet proposes using RSign and RPReLU to enable explicit learning of the distribution reshape and shift at near-zero extra cost.


\subsection{Obtainment: Image Classification}\label{sec:exp_cd}
\textbf{Experimental Setting:} We calculate the $CD$ for several commonly used neural networks, including  ResNet-18~\cite{he2016deep}, MobileNetV2~\cite{howard2017mobilenets}, BNNs (ReActNet and PCNN). The $CD$ is obtained in the image classification task with a SGD optimizer on the CIFAR-10 and ImageNet datasets. In the experiment, we set the learning rate to 0.01, $\alpha=1$ in the Eq.~\ref{eq:u}, and train the network for 256 epochs. We also show the effect of training set size ($10^3$) and training error. We use different numbers of classes in the experiments to explore the effect of various elements on VC-dimension and $CD$. In the section, we will explore the relative generalizations of the different models. The goal is to keep the generalization ranking of different models consistent in different situations. $p$ is calculated based on Eq.~\ref{eq6b}. As shown in Eq.~\ref{eq8}, $p$ is a measure of VC-dimension. We will use the rank of $p$ to represent the rank of VC-dimension.
\begin{table}[t]
\small
\renewcommand\tabcolsep{2pt}
\caption{The $CD$ of different models on variants of CIFAR-10 with batch size of 128. A model of a low $CD$ metric indicated by a high rank has a relatively better generalization ability.}
\label{cifarcd}
\begin{center}
\begin{tabular}{c|ccccc}
\hline
\#Class               & Models      & \begin{tabular}[c]{@{}c@{}}Training set \\ size ($10^3$)\end{tabular} & \begin{tabular}[c]{@{}c@{}}Training \\ error \end{tabular}  & $p$/Rank      & $CD$/Rank       \\ \hline
2  & ReActNet & 10                & 0.512         & 0.385/4 & 0.598/\textbf{4}\\
                    & PCNN        & 10                 & 0.371         & 0.362/3 & 0.582/\textbf{3} \\
                    & ResNet-18   & 10                & 0.034         & 0.102/2 & 0.413/\textbf{2} \\
                    & MobileNetV2   & 10                & 0.004               & 0.022/1       & 0.355/\textbf{1}       \\ \hline
5  & ReActNet & 25                & 0.801         & 0.573/3 & 0.580/\textbf{4} \\
                    & PCNN        & 25                & 0.734         & 0.583/4 & 0.564/\textbf{3} \\
                    & ResNet-18    & 25                & 0.056         & 0.163/2 & 0.275/\textbf{2} \\
                    & MobileNetV2   & 25                &0.001                &0.011/1        &0.229/\textbf{1}        \\ \hline
7  & ReActNet & 35               & 0.785         & 0.638/4 & 0.698/\textbf{4} \\
                    & PCNN        & 35                 & 0.876         & 0.634/3 & 0.696/\textbf{3} \\
                    & ResNet-18    & 35                & 0.124         & 0.291/2 & 0.437/\textbf{2} \\
                    & MobileNetV2   & 35                & 0.001               &  0.010/1      & 0.203/\textbf{1}       \\ \hline
10 & ReActNet & 50                & 0.900  & 0.669/4 & 0.714/\textbf{4} \\
                    & PCNN        & 50                & 0.792         & 0.669/3 & 0.713/\textbf{3} \\
                    & ResNet-18    & 50                & 0.133         & 0.300/2 & 0.423/\textbf{2} \\
                    & MobileNetV2   & 50                & 0.001               &  0.006/1      &  0.272/\textbf{1}      \\ \hline
\end{tabular}
\label{tab:cifar10}
\end{center}
\end{table}

To evaluate the relationship between the generalization ability of BNNs and their full-precision counterparts. We use ResNet-18 as the backbone. To ensure a fair and accurate comparison, we adopt the same experimental setting for different datasets.
The $CD$ of ReActNet and PCNN are shown in Tab.~\ref{tab:cifar10}. We can observe that the $CD$ metric yields a much more stable ranking for these models on various experimental settings, which proves that the $CD$ metric is more stable and reliable than  VC-dimension even when the training set size ($10^3$) and training error vary. 
The $CD$ of PCNN is lower than that of ReActNet, indicating that PCNN has better generalization ability. This explains why the performance of the binary network PCNN is better than  ReActNet in object detection, even though the pre-trained model of ReActNet is much better performance than PCNN.


To further compare the generalization of the $CD$ metric, we also calculate the $CD$ on ImageNet, and the experimental settings are the same with CIFAR-10. The result is shown in  Tab.~\ref{tab:imagenet}, which reveals full-precision networks always have higher generalization than BNNs. Besides, it can be found that the span of $CD$ is much narrower than that of VC-dimension when using different numbers of classes and training set sizes in Tab.~\ref{tab:cifar10} and Tab.~\ref{tab:imagenet}, which might be a good property of $CD$ metric for DNNs. In the conditions of different numbers of classes and different datasets, the ranking of $CD$ is always consistent though the ranking of VC-dimension is disorder. Even the relationship between errors and $CD$ is not monotonic, $CD$ actually provides a complementary measure based on the training error and VC-dimension. As a result, $CD$ is more stable than VC-dimension. 
\begin{table}[t]
\small
\renewcommand\tabcolsep{2pt}
\caption{The $CD$ of different models on variants of ImageNet with batch size of 128.}
\label{imagenetcd}
\begin{center}
\begin{tabular}{c|ccccc}
\hline
\#Class               & Models      & \begin{tabular}[c]{@{}c@{}}Training set \\ size ($10^3$)\end{tabular} & \begin{tabular}[c]{@{}c@{}}Training \\ error \end{tabular}  & $p$/Rank      & $CD$/Rank       \\ \hline
2  & ReActNet & 2                                       & 0.490        & 0.349/3 & 0.579/\textbf{4}                                              \\
                    & PCNN        & 2                                      & 0.322        & 0.354/4 & 0.576/\textbf{3}                                                 \\
                    & ResNet-18   & 2                                                                                    & 0.281        & 0.342/2 & 0.567/\textbf{2}  \\
                    & MobileNetV2   & 2                                                                                    & 0.378       & 0.278/1 & 0.538/\textbf{1}  \\ \hline
10 & ReActNet & 10                                       & 0.117        & 0.645/4 & 0.688/\textbf{4}                                             \\
                    & PCNN        & 10                                      & 0.195        & 0.642/3 & 0.687/\textbf{3}                                               \\
                    & ResNet-18    & 10                                                                                   & 0.082        & 0.601/2 & 0.654/\textbf{2}  \\
                    & MobileNetV2   & 10                                                                                   & 0.180        & 0.574/1 & 0.634/\textbf{1}  \\ \hline
50 & ReActNet & 50                                        &           0.133        & 0.740/4 & 0.752/\textbf{4}                                   \\
                    & PCNN        & 50                                      &  0.087        & 0.701/3 & 0.716/\textbf{3}                                             \\
                    & ResNet-18    & 50                                                                                   & 0.0347        & 0.674/2 & 0.691/\textbf{2} \\
                    & MobileNetV2   & 50                                                                                   & 0.333        & 0.645/1 & 0.667/\textbf{1}  \\ \hline
\end{tabular}
\label{tab:imagenet}
\end{center}
\end{table}

\begin{table}[t]
\small
\renewcommand\tabcolsep{2pt}
\centering
\caption{ReActNet with different optimizers on variants of ImageNet with batch size of 128.}
\begin{tabular}{c|ccccc}
\hline
\#Class               & Models      & \begin{tabular}[c]{@{}c@{}}Training set \\ size ($10^3$)\end{tabular} & \begin{tabular}[c]{@{}c@{}}Training \\ error \end{tabular}  & $p$/Rank      & $CD$/Rank       \\ \hline
2     & SGD     & 2                                                                                  & 0.489        & 0.349/3 & 0.576/\textbf{3}   \\
      & Adam      & 2                      &  0.190	
		
              & 0.212/1 & 0.491/\textbf{2}                                                                \\
      & AdamW     & 2                     &   0.013	
             &  0.228/2 &   0.489/\textbf{1}                                                              \\ \hline
10     & SGD     & 10                                                                                   & 0.117        & 0.645/3 & 0.688/\textbf{3}  \\
      & Adam      & 10                                                                                   &  0.008	
              & 0.328/2	   & 0.441/\textbf{2}   \\
      & AdamW     & 10                                                                                   & 0.004	
               & 0.299/1  & 	0.418/\textbf{1}  \\ \hline
50     & SGD     & 50                                                                                   & 0.134        & 0.740/3 & 0.752/\textbf{3}    \\
      & Adam      & 50                                                                                   &   0.027
             &   	0.658/2	 & 0.677/\textbf{2}   \\
      & AdamW     & 50                                                                                   & 0.019	
               & 0.576/1	   & 0.603/\textbf{1}   \\ \hline

\end{tabular}
\label{tab:optimizer}
\end{table}


We validate the effect of the optimizer for $CD$ in Tab.~\ref{tab:optimizer}. We can find that the volatility with different optimizers is limited to 44\% in $CD$ versus 71\% in VC-dimension. Our $CD$ reserves the same ranking for different parameter settings while VC-dimension is inconsistent, which strongly validates that $CD$ can relatively and stably measure the generalization ability of DNNs. In addition, we  further validate $CD$ on object detection as a stable metric for the generalization ability measure in Section~\ref{sec:exp_detection}.


\subsection{Validation: Object Detection}\label{sec:exp_detection}

\textbf{Experimental setting:} Our experiments are based on Faster R-CNN~\cite{ren2015faster}  for object detection. We test the performance of Faster RCNN with PCNN and ReActNet (backbone is ResNet-18 or MobilNet V2)  in Tab.~\ref{table:Res18voc}, to validate whether the generalization of them are consistent with the rank of the $CD$ as shown in Tab.~\ref{tab:cifar10} and Tab.~\ref{tab:imagenet}. We first initialize the backbone of Faster RCNN from the pre-trained models, \ie, the models from the image classification task. Note that the convolutions in neck and head of Faster RCNN are also binarized by corresponding method. Then, we use the SGD optimizer with a batch size of 8 and a learning rate of 0.0015, to finetune the networks for 36 epochs. 


We evaluate the generalization ability of deep detectors based on the rate of performance change on image classification versus object detection.  Since deep detectors are significantly affected by the initially  pre-trained models based on image classification, we believe the model with lower rate of performance degradation on object detection than that on image classification achieves better generalization. For example, as shown in Table 4, the model,  of larger performance degradation on object classification  ($13.2$\%) but becoming   smaller ($3.9$\%) on object detection, gains a better generalization. Through object detection, we reach a safe conclusion that the generalization ranking of the models is consistent with  $CD$ in Section~\ref{sec:exp_cd}.



\begin{table}[!htp]
\renewcommand\tabcolsep{1pt}
\small
\caption{Test results of BNNs on object detection. The used detection datasets are Pascal VOC and MS COCO. The used classification dataset is ImageNet. `W' and `A' refer to the weight and activation bitwidth, respectively.}
\centering
\begin{tabular}{ccccccc}
\toprule
Backbone     & \begin{tabular}[c]{@{}c@{}}Binary \\ method\end{tabular}  & W  & A  &  \begin{tabular}[c]{@{}c@{}}Accuracy of \\ classification\end{tabular}  &\begin{tabular}[c]{@{}c@{}}  mAP of\\ detection\end{tabular} &\begin{tabular}[c]{@{}c@{}}  Better genera-\\ lized model\end{tabular}   \\ 
\toprule
\multicolumn{7}{c}{VOC2007+ImageNet}                    \\ \hline
ResNet-18    & ReAct       & 1  & 1  & 65.9  &  69.3 &        \\
ResNet-18    & PCNN          & 1  & 1  & 57.2  &  72.0  &   \\ \hline
\multicolumn{4}{c}{Rate of performance change} &$13.2\%$ $\downarrow$  & $3.9\%$ $\uparrow$ & \textbf{PCNN}
\\ \toprule
\multicolumn{7}{c}{VOC2007+ImageNet}                    \\ \hline
ResNet-18    & ReAct       & 1  & 1  & 65.9  &  69.3 & \\
MobileNetV2    & ReAct          & 1  & 1  & 69.5  &74.1    &  \\ \hline
\multicolumn{4}{c}{Rate of performance change} &5.5\% $\uparrow$ & 6.9\%  $\uparrow$  &\textbf{MobileNetV2}
\\ \toprule
\multicolumn{7}{c}{COCO2017+ImageNet}                        \\ \hline
ResNet-18    & ReAct     & 1  & 1    & 65.9  &   27.5  &   \\
ResNet-18    & PCNN          & 1  & 1   & 57.2  &  26.9  &    \\ \hline  
\multicolumn{4}{c}{Rate of performance change} &13.2\% $\downarrow$ & 2.2\% $\downarrow$  &\textbf{PCNN}
\\ \bottomrule
\end{tabular}
	\label{table:Res18voc}
\end{table}

The results of ReActNet and PCNN show that even PCNN achieves a much worse performance on the image classification task than ReActNet, 
it still can achieve close performance (on COCO) or better performance (on VOC) than ReActNet.
These results indicate that PCNN trained on the image classification task has a better generalization than ReActNet.
We also compare ResNet-18, MobileNetV2  with ReAct. We can see that MobileNetV2 achieves a 5.5\%  improvement on the image classification task and a 6.9\%  improvement on the object detection task, which indicates MobileNetV2 performed better on cross-tasks. Conclusively, the generalization of the MobileNetV2 backbone is slightly better than that of the ResNet-18 backbone, which is also in line with the $CD$ ranking. All results demonstrate that the $CD$ is a reliable metric to measure the model's generalization ability.


\section{Conclusion}
\label{sec:conclusion}

In this work, we present a flexible framework to understand the capacity and generalization of DNNs. We introduce  \emph{confidence dimension} ($CD$) to measure deep models' relative generalization ability together with a theoretical analysis, which proves to be the upper bound of the generalization. We validate our  $CD$ on  cross tasks of object recognition and detection over BNNs and their full-precision counterparts. In  future, we will try more applications to validate the effectiveness of our method. 

\bibliographystyle{named}
\bibliography{ijcai22}

\begin{thebibliography}{57}
\providecommand{\natexlab}[1]{#1}
\providecommand{\url}[1]{\texttt{#1}}
\expandafter\ifx\csname urlstyle\endcsname\relax
  \providecommand{\doi}[1]{doi: #1}\else
  \providecommand{\doi}{doi: \begingroup \urlstyle{rm}\Url}\fi

\bibitem[Abelson et~al.(1985)Abelson, Sussman, and
  Sussman]{abelson-et-al:scheme}
Harold Abelson, Gerald~Jay Sussman, and Julie Sussman.
\newblock \emph{Structure and Interpretation of Computer Programs}.
\newblock MIT Press, Cambridge, Massachusetts, 1985.

\bibitem[Arora et~al.(2018)Arora, Ge, Neyshabur, and Zhang]{arora2018stronger}
Sanjeev Arora, Rong Ge, Behnam Neyshabur, and Yi~Zhang.
\newblock Stronger generalization bounds for deep nets via a compression
  approach.
\newblock In \emph{International Conference on Machine Learning}, pages
  254--263. PMLR, 2018.

\bibitem[Bartlett and Mendelson(2002)]{bartlett2002rademacher}
Peter~L Bartlett and Shahar Mendelson.
\newblock Rademacher and gaussian complexities: Risk bounds and structural
  results.
\newblock \emph{The Journal of Machine Learning Research}, 3\penalty0
  (Nov):\penalty0 463--482, 2002.

\bibitem[Baumgartner et~al.(2001)Baumgartner, Gottlob, and Flesca]{bgf:Lixto}
Robert Baumgartner, Georg Gottlob, and Sergio Flesca.
\newblock Visual information extraction with {Lixto}.
\newblock In \emph{Proceedings of the 27th International Conference on Very
  Large Databases}, pages 119--128, Rome, Italy, September 2001. Morgan
  Kaufmann.

\bibitem[Bottou(1998)]{bottou1998online}
L{\'e}on Bottou.
\newblock Online learning and stochastic approximations.
\newblock \emph{On-Line Learning in Neural Networks}, 17\penalty0 (9):\penalty0
  142, 1998.

\bibitem[Bousquet and Elisseeff(2002)]{bousquet2002stability}
Olivier Bousquet and Andr{\'e} Elisseeff.
\newblock Stability and generalization.
\newblock \emph{The Journal of Machine Learning Research}, 2:\penalty0
  499--526, 2002.

\bibitem[Brachman and Schmolze(1985)]{brachman-schmolze:kl-one}
Ronald~J. Brachman and James~G. Schmolze.
\newblock An overview of the {KL-ONE} knowledge representation system.
\newblock \emph{Cognitive Science}, 9\penalty0 (2):\penalty0 171--216,
  April--June 1985.

\bibitem[Courbariaux et~al.(2016)Courbariaux, Hubara, Soudry, El-Yaniv, and
  Bengio]{courbariaux2016binarized}
Matthieu Courbariaux, Itay Hubara, Daniel Soudry, Ran El-Yaniv, and Yoshua
  Bengio.
\newblock Binarized neural networks: Training deep neural networks with weights
  and activations constrained to+ 1 or-1.
\newblock \emph{International Conference on Learning Representations}, 2016.

\bibitem[Deng et~al.(2009)Deng, Dong, Socher, Li, Li, and
  Fei-Fei]{deng2009imagenet}
Jia Deng, Wei Dong, Richard Socher, Li-Jia Li, Kai Li, and Li~Fei-Fei.
\newblock Imagenet: A large-scale hierarchical image database.
\newblock In \emph{Proceedings of the IEEE Conference on Computer Vision and
  Pattern Recognition}, 2009.

\bibitem[Dinh et~al.(2017)Dinh, Pascanu, Bengio, and Bengio]{dinh2017sharp}
Laurent Dinh, Razvan Pascanu, Samy Bengio, and Yoshua Bengio.
\newblock Sharp minima can generalize for deep nets.
\newblock In \emph{International Conference on Machine Learning}, pages
  1019--1028. PMLR, 2017.

\bibitem[Dubhashi and Panconesi(2009)]{dubhashi2009concentration}
Devdatt~P Dubhashi and Alessandro Panconesi.
\newblock \emph{Concentration of measure for the analysis of randomized
  algorithms}.
\newblock Cambridge University Press, 2009.

\bibitem[Dziugaite and Roy(2016)]{dziugaite2017computing}
Gintare~Karolina Dziugaite and Daniel~M Roy.
\newblock Computing nonvacuous generalization bounds for deep (stochastic)
  neural networks with many more parameters than training data.
\newblock In \emph{Thirty-Third Conference on Uncertainty in Artificial
  Intelligence}, 2016.

\bibitem[Dziugaite et~al.(2020)Dziugaite, Drouin, Neal, Rajkumar, Caballero,
  Wang, Mitliagkas, and Roy]{dziugaite2020search}
Gintare~Karolina Dziugaite, Alexandre Drouin, Brady Neal, Nitarshan Rajkumar,
  Ethan Caballero, Linbo Wang, Ioannis Mitliagkas, and Daniel~M Roy.
\newblock In search of robust measures of generalization.
\newblock In \emph{Advances in Neural Information Processing Systems}, 2020.

\bibitem[Everingham et~al.(2015)Everingham, Eslami, Van~Gool, Williams, Winn,
  and Zisserman]{everingham2015pascal}
Mark Everingham, SM~Ali Eslami, Luc Van~Gool, Christopher~KI Williams, John
  Winn, and Andrew Zisserman.
\newblock The pascal visual object classes challenge: A retrospective.
\newblock \emph{International Journal of Computer Vision}, 111\penalty0
  (1):\penalty0 98--136, 2015.

\bibitem[Fort et~al.(2019)Fort, Nowak, Jastrzebski, and
  Narayanan]{fort2019stiffness}
Stanislav Fort, Pawe{\l}~Krzysztof Nowak, Stanislaw Jastrzebski, and Srini
  Narayanan.
\newblock Stiffness: A new perspective on generalization in neural networks.
\newblock In \emph{International Conference on Learning Representations}, 2019.

\bibitem[Girshick et~al.(2014)Girshick, Donahue, Darrell, and
  Malik]{girshick2014rich}
Ross Girshick, Jeff Donahue, Trevor Darrell, and Jitendra Malik.
\newblock Rich feature hierarchies for accurate object detection and semantic
  segmentation.
\newblock In \emph{Proceedings of the IEEE Conference on Computer Vision and
  Pattern Recognition}, 2014.

\bibitem[Gottlob(1992)]{gottlob:nonmon}
Georg Gottlob.
\newblock Complexity results for nonmonotonic logics.
\newblock \emph{Journal of Logic and Computation}, 2\penalty0 (3):\penalty0
  397--425, June 1992.

\bibitem[Gottlob et~al.(2002)Gottlob, Leone, and Scarcello]{gls:hypertrees}
Georg Gottlob, Nicola Leone, and Francesco Scarcello.
\newblock Hypertree decompositions and tractable queries.
\newblock \emph{Journal of Computer and System Sciences}, 64\penalty0
  (3):\penalty0 579--627, May 2002.

\bibitem[Gu et~al.(2019)Gu, Li, Zhang, Han, Cao, Liu, and
  Doermann]{gu2019projection}
Jiaxin Gu, Ce~Li, Baochang Zhang, Jungong Han, Xianbin Cao, Jianzhuang Liu, and
  David Doermann.
\newblock Projection convolutional neural networks for 1-bit cnns via discrete
  back propagation.
\newblock In \emph{Proceedings of the Association for the Advancement of
  Artificial Intelligence Conference on Artificial Intelligence}, 2019.

\bibitem[Han et~al.(2016)Han, Mao, and J.~Dally]{deepcompression}
Song Han, Huizi Mao, and William J.~Dally.
\newblock Deep compression: Compressing deep neural networks with pruning,
  trained quantization and huffman coding.
\newblock In \emph{International Conference on Learning Representations}, 2016.

\bibitem[He et~al.(2016)He, Zhang, Ren, and Sun]{he2016deep}
Kaiming He, Xiangyu Zhang, Shaoqing Ren, and Jian Sun.
\newblock Deep residual learning for image recognition.
\newblock In \emph{Proceedings of the IEEE Conference on Computer Vision and
  Pattern Recognition}, 2016.

\bibitem[Helwegen et~al.(2019)Helwegen, Widdicombe, Geiger, Liu, Cheng, and
  Nusselder]{helwegen2019latent}
Koen Helwegen, James Widdicombe, Lukas Geiger, Zechun Liu, Kwang-Ting Cheng,
  and Roeland Nusselder.
\newblock Latent weights do not exist: Rethinking binarized neural network
  optimization.
\newblock \emph{In Advances in Neural Information Processing Systems}, 2019.

\bibitem[Hou et~al.(2018)Hou, Zhang, and Kwok]{hou2018analysis}
Lu~Hou, Ruiliang Zhang, and James~T Kwok.
\newblock Analysis of quantized models.
\newblock In \emph{International Conference on Learning Representations}, 2018.

\bibitem[Hubara et~al.(2016)Hubara, Courbariaux, Soudry, El-Yaniv, and
  Bengio]{hubara2016binarized}
Itay Hubara, Matthieu Courbariaux, Daniel Soudry, Ran El-Yaniv, and Yoshua
  Bengio.
\newblock Binarized neural networks.
\newblock In \emph{Advances in Neural Information Processing Systems}, 2016.

\bibitem[{IJCAI Proceedings}()]{proceedings}
{IJCAI Proceedings}.
\newblock {IJCAI} camera ready submission.
\newblock \url{https://proceedings.ijcai.org/info}.

\bibitem[Jiang et~al.(2019)Jiang, Neyshabur, Mobahi, Krishnan, and
  Bengio]{jiang2019fantastic}
Yiding Jiang, Behnam Neyshabur, Hossein Mobahi, Dilip Krishnan, and Samy
  Bengio.
\newblock Fantastic generalization measures and where to find them.
\newblock In \emph{International Conference on Learning Representations}, 2019.

\bibitem[Keskar et~al.(2017)Keskar, Nocedal, Tang, Mudigere, and
  Smelyanskiy]{keskar2017large}
Nitish~Shirish Keskar, Jorge Nocedal, Ping Tak~Peter Tang, Dheevatsa Mudigere,
  and Mikhail Smelyanskiy.
\newblock On large-batch training for deep learning: Generalization gap and
  sharp minima.
\newblock In \emph{International Conference on Learning Representations}, 2017.

\bibitem[Kingma and Ba(2015)]{kingma2014adam}
Diederik~P Kingma and Jimmy Ba.
\newblock Adam: A method for stochastic optimization.
\newblock In \emph{International Conference on Learning Representations}, 2015.

\bibitem[Krizhevsky and Hinton(2009)]{krizhevsky2009learning}
Alex Krizhevsky and Geoffrey Hinton.
\newblock Learning multiple layers of features from tiny images.
\newblock \emph{Technical report, University of Toronto}, 2009.

\bibitem[Levesque(1984{\natexlab{a}})]{levesque:belief}
Hector~J. Levesque.
\newblock A logic of implicit and explicit belief.
\newblock In \emph{Proceedings of the Fourth National Conference on Artificial
  Intelligence}, pages 198--202, Austin, Texas, August 1984{\natexlab{a}}.
  American Association for Artificial Intelligence.

\bibitem[Levesque(1984{\natexlab{b}})]{levesque:functional-foundations}
Hector~J. Levesque.
\newblock Foundations of a functional approach to knowledge representation.
\newblock \emph{Artificial Intelligence}, 23\penalty0 (2):\penalty0 155--212,
  July 1984{\natexlab{b}}.

\bibitem[Li et~al.(2019{\natexlab{a}})Li, Luo, and Qiao]{li2019generalization}
Jian Li, Xuanyuan Luo, and Mingda Qiao.
\newblock On generalization error bounds of noisy gradient methods for
  non-convex learning.
\newblock In \emph{International Conference on Learning Representations},
  2019{\natexlab{a}}.

\bibitem[Li et~al.(2019{\natexlab{b}})Li, Wang, Liang, Qin, Yan, and
  Fan]{li2019fully}
Rundong Li, Yan Wang, Feng Liang, Hongwei Qin, Junjie Yan, and Rui Fan.
\newblock Fully quantized network for object detection.
\newblock In \emph{Proceedings of the IEEE Conference on Computer Vision and
  Pattern Recognition}, pages 2810--2819, 2019{\natexlab{b}}.

\bibitem[Lin et~al.(2014)Lin, Maire, Belongie, Hays, Perona, Ramanan,
  Doll{\'a}r, and Zitnick]{lin2014microsoft}
Tsung-Yi Lin, Michael Maire, Serge Belongie, James Hays, Pietro Perona, Deva
  Ramanan, Piotr Doll{\'a}r, and C~Lawrence Zitnick.
\newblock Microsoft coco: Common objects in context.
\newblock In \emph{Proceedings of the European Conference on Computer Vision},
  2014.

\bibitem[Liu et~al.(2019)Liu, Bai, Jiang, Chen, and Wang]{liu2019understanding}
Jinlong Liu, Yunzhi Bai, Guoqing Jiang, Ting Chen, and Huayan Wang.
\newblock Understanding why neural networks generalize well through gsnr of
  parameters.
\newblock In \emph{International Conference on Learning Representations
  (ICLR)}, 2019.

\bibitem[Liu et~al.(2018)Liu, Wu, Luo, Yang, Liu, and Cheng]{liu2018bi}
Zechun Liu, Baoyuan Wu, Wenhan Luo, Xin Yang, Wei Liu, and Kwang-Ting Cheng.
\newblock Bi-real net: Enhancing the performance of 1-bit cnns with improved
  representational capability and advanced training algorithm.
\newblock In \emph{Proceedings of the European Conference on Computer Vision},
  2018.

\bibitem[Liu et~al.(2020)Liu, Shen, Savvides, and Cheng]{liu2020reactnet}
Zechun Liu, Zhiqiang Shen, Marios Savvides, and Kwang-Ting Cheng.
\newblock Reactnet: Towards precise binary neural network with generalized
  activation functions.
\newblock In \emph{Proceedings of the European Conference on Computer Vision},
  2020.

\bibitem[Loshchilov and Hutter(2018)]{loshchilov2018fixing}
Ilya Loshchilov and Frank Hutter.
\newblock Fixing weight decay regularization in adam.
\newblock In \emph{International Conference on Learning Representations}, 2018.

\bibitem[Mukherjee et~al.(2004)Mukherjee, Niyogi, Poggio, and
  Rifkin]{mukherjee2004statistical}
Sayan Mukherjee, Partha Niyogi, Tomaso Poggio, and Ryan Rifkin.
\newblock Statistical learning: Stability is sufficient for generalization and
  necessary and sufficient for consistency of empirical risk minimization.
\newblock Technical report, 2004.

\bibitem[Nagarajan and Kolter(2019)]{nagarajan2019uniform}
Vaishnavh Nagarajan and J~Zico Kolter.
\newblock Uniform convergence may be unable to explain generalization in deep
  learning.
\newblock \emph{In Advances in Neural Information Processing Systems}, 2019.

\bibitem[Nebel(2000)]{nebel:jair-2000}
Bernhard Nebel.
\newblock On the compilability and expressive power of propositional planning
  formalisms.
\newblock \emph{Journal of Artificial Intelligence Research}, 12:\penalty0
  271--315, 2000.

\bibitem[Neyshabur et~al.(2017)Neyshabur, Bhojanapalli, McAllester, and
  Srebro]{neyshabur2017exploring}
Behnam Neyshabur, Srinadh Bhojanapalli, David McAllester, and Nathan Srebro.
\newblock Exploring generalization in deep learning.
\newblock In \emph{Advances in Neural Information Processing Systems}, 2017.

\bibitem[Neyshabur et~al.(2019)Neyshabur, Li, Bhojanapalli, LeCun, and
  Srebro]{neyshabur2018towards}
Behnam Neyshabur, Zhiyuan Li, Srinadh Bhojanapalli, Yann LeCun, and Nathan
  Srebro.
\newblock Towards understanding the role of over-parametrization in
  generalization of neural networks.
\newblock In \emph{International Conference on Learning Representations}, 2019.

\bibitem[Poggio et~al.(2004)Poggio, Rifkin, Mukherjee, and
  Niyogi]{poggio2004general}
Tomaso Poggio, Ryan Rifkin, Sayan Mukherjee, and Partha Niyogi.
\newblock General conditions for predictivity in learning theory.
\newblock \emph{Nature}, 428\penalty0 (6981):\penalty0 419--422, 2004.

\bibitem[Rastegari et~al.(2016)Rastegari, Ordonez, Redmon, and
  Farhadi]{paper09}
Mohammad Rastegari, Vicente Ordonez, Joseph Redmon, and Ali Farhadi.
\newblock Xnor-net: Imagenet classification using binary convolutional neural
  networks.
\newblock In \emph{Proceedings of the European Conference on Computer Vision},
  2016.

\bibitem[Ren et~al.(2015)Ren, He, Girshick, and Sun]{ren2015faster}
Shaoqing Ren, Kaiming He, Ross~B Girshick, and Jian Sun.
\newblock Faster r-cnn: Towards real-time object detection with region proposal
  networks.
\newblock In \emph{Advances in Neural Information Processing Systems}, 2015.

\bibitem[Sandler et~al.(2018)Sandler, Howard, Zhu, Zhmoginov, and
  Chen]{howard2017mobilenets}
Mark Sandler, Andrew Howard, Menglong Zhu, Andrey Zhmoginov, and Liang-Chieh
  Chen.
\newblock Mobilenetv2: Inverted residuals and linear bottlenecks.
\newblock In \emph{Proceedings of the IEEE conference on computer vision and
  pattern recognition}, 2018.

\bibitem[Srivastava et~al.(2014)Srivastava, Hinton, Krizhevsky, Sutskever, and
  Salakhutdinov]{srivastava2014dropout}
Nitish Srivastava, Geoffrey Hinton, Alex Krizhevsky, Ilya Sutskever, and Ruslan
  Salakhutdinov.
\newblock Dropout: a simple way to prevent neural networks from overfitting.
\newblock \emph{The Journal of Machine Learning Research}, 15\penalty0
  (1):\penalty0 1929--1958, 2014.

\bibitem[Sun et~al.(2016)Sun, Chen, Wang, Liu, and Liu]{sun2016depth}
Shizhao Sun, Wei Chen, Liwei Wang, Xiaoguang Liu, and Tie-Yan Liu.
\newblock On the depth of deep neural networks: A theoretical view.
\newblock In \emph{Proceedings of the Association for the Advance of Artificial
  Intelligence Conference on Artificial Intelligence}, 2016.

\bibitem[Valle-P{\'e}rez and Louis(2020)]{valle2020generalization}
Guillermo Valle-P{\'e}rez and Ard~A Louis.
\newblock Generalization bounds for deep learning.
\newblock In \emph{International Conference on Learning Representations}, 2020.

\bibitem[Valle-Perez et~al.(2018)Valle-Perez, Camargo, and
  Louis]{valle2018deep}
Guillermo Valle-Perez, Chico~Q Camargo, and Ard~A Louis.
\newblock Deep learning generalizes because the parameter-function map is
  biased towards simple functions.
\newblock In \emph{International Conference on Learning Representations}, 2018.

\bibitem[Vapnik and Chervonenkis(1971)]{vapnik1971uniform}
VN~Vapnik and A~Ya Chervonenkis.
\newblock On the uniform convergence of relative frequencies of events to their
  probabilities.
\newblock \emph{Measures of Complexity}, 16\penalty0 (2):\penalty0 11, 1971.

\bibitem[Wu et~al.(2018)Wu, Li, Chen, and Shi]{ICLR2018wu}
Shuang Wu, Guoqi Li, Feng Chen, and Luping Shi.
\newblock {Training and inference with integers in deep neural networks}.
\newblock In \emph{International Conference on Learning Representations}, 2018.

\bibitem[Yin et~al.(2019)Yin, Lyu, Zhang, Osher, Qi, and
  Xin]{yin2019understanding}
P~Yin, J~Lyu, S~Zhang, S~Osher, YY~Qi, and J~Xin.
\newblock Understanding straight-through estimator in training activation
  quantized neural nets.
\newblock In \emph{International Conference on Learning Representations}, 2019.

\bibitem[Zhang et~al.(2017)Zhang, Bengio, Hardt, Recht, and
  Vinyals]{zhang2016understanding}
Chiyuan Zhang, Samy Bengio, Moritz Hardt, Benjamin Recht, and Oriol Vinyals.
\newblock Understanding deep learning requires rethinking generalization.
\newblock In \emph{International Conference for Learning Representations},
  2017.

\bibitem[Zhou et~al.(2017)Zhou, Ye, Qiu, and Jiao]{CVPR2018zhou}
Yanzhao Zhou, Qixiang Ye, Qiang Qiu, and Jianbin Jiao.
\newblock Oriented response networks.
\newblock In \emph{IEEE Conference on Computer Vision and Pattern Recognition},
  2017.

\bibitem[Zoph and Le(2016)]{zoph2016neural}
B.~Zoph and Q.~V. Le.
\newblock Neural architecture search with reinforcement learning.
\newblock In \emph{International Conference on Learning Representations}, 2016.

\end{thebibliography}

\end{document}